# Reproducible Benchmarking for Lung Nodule Detection and Malignancy Classification Across Multiple Low-Dose CT Datasets

**Original Research**

Fakrul Islam Tushar MS[1,2], Avivah Wang MD[3], Lavsen Dahal MS[1,2], Ehsan Samei PhD[1,2,3], Michael R. Harowicz MD[4], Jayashree Kalpathy-Cramer PhD[5], Kyle J. Lafata PhD[1,2], Tina D. Tailor MD[4], Cynthia Rudin PhD[6], Joseph Y. Lo PhD[1,2,3]

[1] Center for Virtual Imaging Trials, Carl E. Ravin Advanced Imaging Laboratories, Department of Radiology, Duke University School of Medicine, Durham, NC

[2] Dept. of Electrical & Computer Engineering, Pratt School of Engineering, Duke University, Durham

[3] Duke University School of Medicine, Durham, NC

[4] Division of Cardiothoracic Imaging, Department of Radiology, Duke University School of Medicine, Durham, NC

[5] Dept. of Ophthalmology, University of Colorado, Boulder, Colorado

[6] Dept. of Computer Science, Duke University, Durham

## Abstract

**Background:** Evaluation of artificial intelligence (AI) models for low-dose CT lung cancer screening is limited by heterogeneous datasets and annotation standards, making performance difficult to compare and translate across clinical settings.

**Purpose:** To establish a public, reproducible multi-dataset benchmark for lung nodule detection and nodule-level cancer classification and to quantify cross-dataset generalizability.

**Materials & Methods:** This retrospective study used Duke Lung Cancer Screening (DLCS), a large and well-annotated dataset, to develop models and to compare performances on three other datasets: LUNA16/LIDC-IDRI, NLST-3D, and LUNA25. For the first task, detection models were trained on DLCS and LUNA16 and evaluated using free-response ROC externally on NLST-3D. For the second task of nodule-level cancer classification, we compared five model types: randomly initialized ResNet50, Models Genesis, Med3D, Foundation Model for Cancer Biomarkers, and Strategic Warm-Start (ResNet50-SWS) pretrained with detection-derived candidate patches stratified by confidence. Classification performance was summarized by AUC with 95% confidence intervals and DeLong tests.

**Results:** Detection model performance varied across datasets, with training on clinically curated annotations (DLCS) outperforming training on research-focused annotations (LUNA16), achieving higher sensitivity at 2 FP/scan on external validation with NLST-3D (0.72 vs 0.64; $p < .001$). For malignancy classification, performance also differed substantially by dataset, with ResNet50-SWS achieving AUCs of 0.71 (DLCS; 95% CI, 0.61-0.81), 0.90 (LUNA16; 0.87-0.93), 0.81 (NLST-3D; 0.79-0.82), and 0.80 (LUNA25; 0.78-0.82), matching or exceeding the other four classification strategies. ResNet50-SWS significantly outperformed randomly initialized ResNet50 model and Models Genesis on all large external datasets ($p < .001$).

**Conclusion:** This study establishes a transparent, multi-dataset benchmark that demonstrates lung cancer detection and classification performance is strongly driven by dataset characteristics. This benchmark framework provides reproducible evaluation of lung nodule AI under differing reference standards, supporting informed comparison and future translational studies.

**Summary Statement**

To evaluate low-dose CT lung nodule AI, we curated a reproducible, multi-dataset benchmark to demonstrate that detection and malignancy classification performance varies substantially by dataset and reference standard.

**Key Results**

- We publish a reproducible, multi-dataset benchmark for CT lung nodule detection and malignancy classification, enabling transparent comparisons.
- Training on clinically curated annotations (DLCS) improved external detection performance on NLST-3D compared with research-focused annotations (LUNA16) (sensitivity 0.72 vs 0.64 at 2 false-positives/scan; $p < 0.001$).
- Classification performance varied substantially by dataset and annotation standards, with detection-informed pretraining performing comparably to common pretrained/self-supervised baselines across datasets (AUC: DLCS 0.71; LUNA16 0.90; NLST-3D 0.81; LUNA25 0.80).

# 1. Introduction

Low-dose chest CT is the primary imaging modality for lung cancer screening (1, 2). Radiologist interpretation of CT exams is time-consuming, subject to observer variability, and challenged by subtle findings and high false-positive rates (2, 3). Artificial intelligence (AI), particularly advances in deep learning, may assist radiologists by improving performance and reducing workload. Realizing that potential requires not only large, high-quality datasets but also reproducible benchmarking frameworks to support both rigorous training and reproducible evaluation.

Lung nodule detection and malignancy classification research has relied on public datasets, including National Lung Screening Trial (NLST) (1), Lung Image Database Consortium and Image Database Resource Initiative (LIDC-IDRI) (4, 5), LUNA16 (6), and LUNA25 (7), often supplemented by private data or additional annotations (8). These datasets differ markedly in dataset sizes, annotation granularity, and reference standards. For example, NLST comprises 26,722 participants but only 1,060 cancer patients due to the low prevalence and no lesion annotations (1). The Sybil study annotated nodules from NLST cancer patients only (9), and the recent LUNA25 challenge dataset was also derived from the NLST (7). Derived from LIDC-IDRI, LUNA16 contained >1,100 lesion annotations from >600 CT scans, with subjective radiologist malignancy scores as a proxy reference standard (8, 10).

These datasets spurred many deep learning approaches. For nodule detection, the LUNA16 challenge established a standardized evaluation protocol (6) and motivated many convolutional neural network approaches (11), including the nnDetection self-configuring framework (12). MONAI’s open-source RetinaNet (13) implementation employed similar self-configuring workflows (14). For nodule classification, several new approaches have emerged for this data-limited scenario. Med3D was pretrained on eight public 3D medical imaging segmentation datasets (15) and Models Genesis used self-supervised pretraining (10). The SimCLR variant Foundation Model for Cancer Biomarkers (FMCB) was trained on over 11,000 CT lesions, then those features were used to train a regression classifier on LUNA16 to estimate radiologist suspicion scores. Related studies used NLST data to predict the risk of future cancer diagnosis (9, 16).

Despite these methodological advances, reliance on a small number of heterogeneous datasets continues to limit generalizability. Addressing this gap, we leveraged the recently published Duke Lung Cancer Screening (DLCS) dataset with >1600 low-dose CTs with >2400 lesion annotations and linked pathology to systematically train and benchmark a range of nodule detection and classification approaches using multiple datasets, including DLCS, LUNA16, NLST, and LUNA25. As illustrated in Figure 1, this benchmark explicitly separates nodule detection and cancer classification tasks, applies a consistent training and evaluation protocol, and incorporates datasets with differing reference standards, including histopathology and radiologist annotation. This design enables transparent assessment of cross-dataset generalization. All code, pretrained models, and experimental configurations are publicly released as a benchmarking resource.

# 2. Methods

## 2.1. Datasets

We utilized the DLCS (17) as the primary development dataset. For detection, we trained the same model separately on DLCS and LUNA16 (6), and external validation was performed on NLST-3D (1, 9). For classification, all model architectures were trained exclusively on DLCS and externally validated on LIDC-IDRI (5), LUNA16, LUNA25, and NLST-3D datasets. The dataset relationships, reference standards, and task-specific usage are visually summarized in **Figure 1**. Train-test splits were randomly stratified by labels; full details including patient/exam numbers are in **Supplementary Table S1.** In brief, there were notable annotation differences across the datasets. DLCS and the NLST-derived LUNA25 include histopathological confirmed cancers, and all nodules (both cancer and non-cancer) were annotated manually. NLST-3D is also derived from NLST and contains manually annotated cancers but pseudo-labeled non-cancer candidates. Finally, the IDRI-derived LUNA16 relies on radiologist suspicion labels (RSLs) assigned subjectively without histopathology confirmation. Only DLCS provides Lung-RADS categories.

**Duke Lung Cancer Screening (DLCS) Dataset:** This dataset includes 1,613 patients and 2,487 nodules from Duke Health, each marked with a 3D bounding box and clinical and pathological outcomes (17, 18). The initial annotation phase employed MONAI RetinaNet to identify nodule candidates (19), which were verified by a medical student supervised by cardiothoracic imaging radiologists (17). This process adhered to Lung-RADS v2022 (20) by focusing on nodules ≥4mm or located in central or segmental airways. We used 88% of the publicly available data as the development set (training and validation) and reported performance over the reserved 12% test set (**Supplementary Table S1**). Patient demographics and data statistics are detailed in Table 1. This public data is available at Zenodo: 10.5281/zenodo.13799069.

**LIDC-IDRI and LUNA16 Dataset:** Derived from the LIDC-IDRI (5) dataset, LUNA16 includes 601 CT scans with 1,186 annotated nodules. Detection performance was reported based on their predefined 10-fold cross-validation protocol. Since these annotations lack confirmed outcomes, we adopted a proxy standard from a prior study (8) for 677 nodules if at least one LIDC-IDRI radiologist provided a high malignancy score, hereafter the Radiologist Suspicion Label (RSL) (**Fig. 1**).

**National Lung Screening Trial (NLST), LUNA25, and NLST-3D Datasets:**

NLST is the largest and most widely recognized resource for CT-based lung cancer screening. We incorporate two NLST variant datasets for external validation: LUNA25 and NLST-3D. Recently released, **LUNA25** added annotations for 6,163 nodules across 4,069 CT scans from 2,120 patients (7). Each annotated nodule includes 3D lesion center coordinates (x,y,z), sex, and age. Nodules were annotated by a radiologist and two medical students, and nodule-level malignancy labels were based on NLST patient-level labels. LUNA25 was used for external validation of lung cancer classification.

**NLST-3D** was adapted from the Sybil dataset (9), in which radiologists re-annotated over 9,000 2D slice-level bounding boxes from >900 NLST cancer patients. To obtain 3D nodule annotations, we grouped slice-level boxes belonging to the same nodule and into the smallest enclosing 3D box, yielding **1,192** positive annotations in NLST cancer patients. To add benign nodule annotations, we applied the DLCS detection model (**Section 2.2.1**) to patients and selected the two highest-confidence negative candidates per CT (median score, 0.98), yielding 1,936 non-cancer candidates (**Fig. 1**). This enables direct comparison between pseudo-labeled negatives and true benign nodules (e.g., LUNA25). **Table 1 and Supplement Table S1** detailed the study dataset.

## 2.2. Benchmark Tasks

All models were trained and evaluated using a standardized preprocessing (spatial resampling to $0.7 \times 0.7 \times 1.25\ mm$, HU windowing -1000 to 500) and consistent labeling rules established during dataset curation (**Figure 1** and **Supplementary Table S1**), ensuring that inputs from different datasets were comparable.

### 2.2.1. Lung Nodule Detection

The detection task requires locating lung nodules in CT and producing 3D bounding boxes.

**Model Development**. We trained 3D RetinaNet detection models using the MONAI detection workflow (6, 13, 14). The primary model, DLCS-De ("De" for detection), was trained on the DLCS development set with 22% withheld for validation to select checkpoints. To demonstrate the effect of training datasets, we trained LUNA16-De with the LUNA16 10-fold cross-validation protocol (14). For external evaluations, we use the median-performing fold six model to represent the cross-validations. Preprocessing included resampling volumes to $0.7 \times 0.7 \times 1.25\ mm$ and Hounsfield Unit clipping (-1000 to 500) with standardization. The models utilized patch sizes of $192 \times 192 \times 80\ (x, y, z)$ and employed sliding window outputs. Models were trained with identical hyperparameters and training epochs.

**Evaluation**. The DLCS-De model was evaluated on the DLCS test dataset and externally validated on LUNA16. The LUNA16-De model performance was the test result of the median-performing split from the 10-fold cross-validation. Both models were also externally validated on the NLST-3D dataset. Performance was assessed by free-response receiver operating characteristic (FROC) analysis and the LUNA16 Competition Performance Metric (CPM) (6), defined as average sensitivity at 1/8, 1/4, 1/2, 1, 2, 4, and 8 false positives (FP) per scan (6, 21). Sensitivity was also reported at 2 FP/scan to reflect a single, more pragmatic operating point. The LUNA16 protocol applies an exclusion list to omit certain candidates from evaluation (6). DLCS and NLST-3D evaluations do not employ such exclusions to reflect a more clinically representative case mix.

### 2.2.2. Lung Cancer Classification Task

The classification task labels each nodule candidate as cancer or non-cancer. Standardized preprocessing used for the detection task was applied. Nodule-centered $64 \times 64 \times 64$ voxel patches were extracted. All models were trained for 200 epochs, and the final model was selected based on best validation performance.

**Model Development.** Five approaches were trained on the DLCS development set and used to classify each patch:

1) 3D ResNet50 with randomly initialized weights (22).
2) FMCB+: This variant of the Foundation Model for Cancer imaging Biomarkers (FMCB) (8), a self-supervised 3D ResNet50, was trained and then used as a feature extractor. Using 4,096 features per patch, we trained a logistic regression model for classification.
3) Models Genesis (10) Chest CT 3D pretrained model was appended with a classification layer and end-to-end fine-tuned.
4) Med3D ResNet50 (15) pretrained model was similarly end-to-end fine-tuned.
5) ResNet50-SWS: We proposed Strategic Warm-Start (SWS), which strategically uses pretraining to initialize the classifier with task-relevant weights learned from an established, related task, reducing the amount of labeled malignancy data needed and improving false-positive suppression. This approach follows three stages (**Figure 2**).
    a. Candidate regions were extracted from the DLCS-De detection outputs. Positive patches contained annotated nodules, while negative samples were stratified equally by detection confidence scores into three bins: [0%, 40%), [40%, 70%), and [70%, 100%]. Negative samples were intentionally overrepresented at a 3:1 ratio relative to the positive class to encourage false positive suppression.
    b. ResNet50 model with randomly initialized weights was pretrained to classify these selected candidates as nodule or non-nodule, enabling the network to learn relevant lung anatomy and nodule characteristics.

c. Pretrained weights from this candidate classifier were transferred to initialize a downstream malignancy classifier, which was then end-to-end fine-tuned to differentiate malignant from benign nodules.

**Evaluation**. Nodule-level cancer classification performance was evaluated on the DLCS internal test set, then externally validated on LUNA16, LUNA25, and NLST-3D. Performance was assessed using the receiver operating characteristic (ROC) area under the curve (AUC) with 95% confidence intervals (CIs) (23).

# 3. Results

**Supplement Table S1** and **Table 1** displays the number of patients and volumes utilized in model development and testing. The average age of patients in the test datasets was 66 years (range: 54 to 79) for DLCS, 62 years (55 to 76) for LUNA25, and 63 years (55 to 74) for NLST-3D. Males comprised 42%, 57%, and 59% of the DLCS, LUNA25, and NLST-3D test datasets, respectively. No exclusions were made based on age, scanner equipment, protocols, or type of reconstruction.

## 3.1. Nodule Detection

**Benchmark performance on the LUNA16 challenge.** To contextualize detector performance against the established LUNA16 challenge, we report results using the official LUNA16 challenge evaluation protocol at the predefined false-positive (FP) per scan operating points (1/8–8 FP/scan), along with the competition performance metric (CPM), defined as the mean sensitivity across these operating points. **Table 2** presents the official benchmark results: DLCS-De (ours) achieved sensitivities of 0.80, 0.86, 0.91, 0.94, 0.97, 0.98, and 0.99 at 1/8, 1/4, 1/2, 1, 2, 4, and 8 FP/scan, respectively (CPM 0.92), comparable to published methods (Liu et al. (11), CPM 0.92; nnDetection (12), CPM 0.93) and within 0.02 CPM of LUNA16-De (CPM 0.94). DLCS-De testing matched the cross-validation performance of LUNA16-De and nnDetection with a sensitivity of 0.97 at 2 FP/scan, followed by 0.94 for both Liu et al. (11, 12). Notably, published LUNA16 benchmark results are typically reported under the predefined ten-fold cross-validation protocol, whereas our DLCS-De performance represents fully external testing on LUNA16.

**Internal testing on DLCS.** On the DLCS internal test dataset (198 scans in common for paired evaluation), the DLCS-trained detection model (DLCS-De) outperformed the LUNA16-trained detection model (LUNA16-De) across the FROC operating range (**Fig. 3a**). Average sensitivity (0.125-8 FP/scan) increased from 0.57 (95% CI: 0.53-0.62) for LUNA16-De to 0.64 (0.59-0.68) for DLCS-De, corresponding to a paired improvement of 0.061 (95% CI: 0.03-0.09; P < .001). At 2 FP/scan, sensitivity improved from 0.72 (0.67-0.78) to 0.82 (0.76-0.86), yielding a paired gain of 0.09 (0.04-0.14; P < .001).

**On the external NLST-3D test dataset**, the DLCS-De outperformed the LUNA16-De across the FROC operating range (**Fig. 3b**). Average sensitivity (1/8-8 FP/scan) increased from 0.49 (95% CI: 0.47-0.52) for LUNA16-De to 0.58 (0.56-0.61) for DLCS-De. Using a paired bootstrap comparison on the 969 scans common to both evaluations, DLCS-De achieved a mean absolute improvement of 0.093 (95% CI: 0.08-0.11; p < .001) in average sensitivity. At the clinically relevant operating point of 2 FPs/scan, sensitivity improved from 0.64 (0.60-0.67) to 0.72 (0.69-0.75), corresponding to a paired gain of 0.083 (0.06-0.11; < .001). Paired bootstrap comparisons for average sensitivity and sensitivity at 2 FP/scan across DLCS (internal) and NLST-3D (external) are summarized in **Supplementary Table S2**.

### 3.2. Lung Cancer Classification

**Table 3** summarizes nodule-level malignancy classification performance across four evaluation datasets, reporting bootstrapped mean AUCs (95% CIs) and statistical significance versus the reference model (ResNet50-SWS) using the DeLong test. **Figure 4** presents the corresponding ROC curves and AUC comparisons across datasets.

On the DLCS internal test dataset (n=294; 33 malignant), ResNet50-SWS achieved an AUC of 0.71 (0.61-0.81), with FMBI showing comparable discrimination (0.71; 95% CI: 0.60-0.82) and Med3D (0.67; 0.57-0.77) and Genesis (0.64; 0.53-0.75) performing lower; the randomly initialized ResNet50 baseline yielded 0.60 (0.49-0.70).

In external validation on LUNA16 (n=677), ResNet50-SWS achieved the highest AUC (0.90; 0.87-0.93), exceeding FMBI (0.87; 0.84-0.90; p < 0.05), while ResNet50 (0.78; 0.74-0.82; p < 0.001), Genesis (0.78; 0.74-0.81; p < 0.001), and Med3D (0.78; 0.75-0.82; p < 0.001) were significantly lower.

On the large screening dataset NLST-3D (n=3128), ResNet50-SWS again led (0.81 [0.79-0.82]), with FMBI slightly reduced (0.79; 0.77-0.80; p < 0.05) and ResNet50 (0.63 [0.61–0.65]; p < 0.001), Genesis (0.51; 0.48-0.53]; p < 0.001), and Med3D (0.74; 0.72-0.76; p < 0.001) significantly underperforming.

Finally, on LUNA25 (n=6163), ResNet50-SWS achieved 0.80 (0.78-0.82), while FMBI achieved the highest AUC (0.82; 0.80-0.83) and Med3D remained comparable (0.80; 0.78-0.82); in contrast, Genesis (0.51; 0.49-0.54; p < 0.001) and ResNet50 (0.75; 0.73-0.78; p < 0.001) were significantly lower. Collectively, these results indicate that the proposed Strategic Warm-Start pretraining yields consistently strong discrimination across heterogeneous label standards, with generalization across datasets (**Figure 4;**

**Table 3**). **Figure 5** shows examples of cancer/non-cancer 3D sub-volume patches and associated model outputs.

# 4. Discussion

Variability in dataset quality and annotation standards continues to challenge model generalizability and reproducibility (24). The objective of this study was to assemble several large, well-annotated public datasets and create a benchmarking framework for fair comparison and evaluation of CT-based lung cancer AI. By leveraging the DLCS dataset together with the LUNA16, LUNA25, and NLST-3D datasets, we systematically evaluated MONAI RetinaNet-based lung nodule detection models. We also compared five nodule-level cancer classification strategies, including our Strategic Warm-Start (SWS) approach that uses detection-informed pretraining to enhance the downstream classification. Using consistent preprocessing, training, and evaluation protocols, we sought to provide fair comparisons that show how dataset mix, annotation standards, and modeling framework affect performance and generalizability.

For lung nodule detection, the DLCS-trained detector (DLCS-De) achieved higher sensitivity and CPM than the model trained on LUNA16 when externally validated on the NLST-3D. When externally validated on the LUNA16 benchmark, the DLCS-De model matched top LUNA16 internal cross-validation performances (6). Despite the differences among datasets, both models adapted well when applied to the NLST-3D datasets, suggesting a level of transferability that could be beneficial in real-world clinical scenarios. Dataset curation and evaluation rules influenced performances: LUNA16's exclusion protocol focuses evaluation on more obvious, high-concordance nodules, which elevates sensitivity relative to evaluating all annotations. Since the hyperparameter choices were fixed for both DLCS-De and LUNA16-De models, performance differences likely reflect dataset curation or evaluation criteria rather than intrinsic model superiority, underscoring the importance of benchmarking context.

For nodule-level classification, all models were developed on DLCS and externally validated across three datasets: LUNA16, LUNA25, and NLST-3D. Performance varied substantially with the reference standard and case mix. When evaluated against the LUNA16 radiologist suspicion labels, all models showed high AUCs, but those labels lack histopathologic diagnoses and therefore lead to performance that is overestimated. While our results remain competitive with prior studies (8, 25, 26), models based on such subjective labels should be interpreted cautiously. Similarly, LUNA25 and NLST-3D dataset both included pseudo-labeled negatives (from medical students and a detection model, respectively), which were easier to classify and elevated performances. That said, the similarity between these LUNA25 and NLST-3D results suggest that, when pathology is unavailable, pseudo-labeling can still be practically useful. In contrast, performance was notably lower on DLCS because it was curated to include actionable nodules and exclude obvious negatives, thus deliberately concentrating on the challenging task of discriminating suspicious nodules. Models that perform well on a clinically focused, harder dataset such as DLCS may have greater potential for translational relevance.

Prior work suggested the value of large-scale pretraining (15) and self-supervised learning methods (8, 10). By leveraging task-relevant supervised pretraining from the detection pipeline, our Strategic Warm-Start (SWS) classifier matched or exceeded those other pretraining approaches (Models Genesis, Med3D, and FMCB) across external validations. By focusing pretraining on hard negatives and a representative distribution of candidates, SWS accelerated learning of nodule features and transferred effectively to malignancy classification while maintaining the same network architecture and development dataset. When large external pretraining datasets are unavailable or unsuitable due to the domain distribution, SWS appears to provide an alternative that is effective and practical.

This study had limitations. Although DLCS has a large number of high-quality annotations compared with existing public datasets, this single center may underrepresent scanner, protocol, and population heterogeneity, which can limit generalizability. All the datasets in this study remain modest relative to the requirements to train large-scale models (27). To meet the demands for large, diverse training data and reduce reliance on manual annotation, there are increasingly alternative approaches such as biology-informed simulation (28) and diffusion-based generative synthesis (29). Reflecting common patterns in the literature, several external validations rely on proxy or pseudo-labels, which introduce label noise and potential bias. Incorporating prior imaging has been shown in other studies to improve performance (30); however, longitudinal exams are not yet widely available, so many lung nodule AI studies still rely on imaging from a single timepoint, supporting the need for future study. Finally, this work focuses on retrospective, nodule-level evaluation, whereas prospective, multi-center validation with patient-level assessment are needed before clinical deployment.

In conclusion, we demonstrate the feasibility and utility of a public, reproducible multi-dataset benchmark for CT lung nodule detection and malignancy classification. Our results show that apparent performance is strongly influenced by dataset composition and reference standards, underscoring the need to report evaluations across multiple datasets under consistent protocols. By releasing curated datasets, code, and pretrained baselines, we enable transparent comparisons and facilitate external validation and future extensions.

## Acknowledgments

This work was supported by the Center for Virtual Imaging Trials, NIH/NIBIB P41 EB028744, NIH/NIBIB R01 EB038719, and the Putman Vision Award awarded by the Department of Radiology of Duke University School of the Medicine. Data was derived from the Duke Lung Cancer Screening Program

## Data and Code Availability

We have publicly released all code, pretrained models, and baseline results associated with this study. These resources are available at the following repositories:

**GitLab: https://gitlab.oit.duke.edu/cvit-public/ai_lung_health_benchmarking**

**GitHub: https://github.com/fitushar/AI-in-Lung-Health-Benchmarking-Detection-and-DiagnosticModels-Across-Multiple-CT-Scan-Datasets**

The **Duke Lung Cancer Screening (DLCS) dataset**, including diagnostic labels and bounding box annotations, is publicly available via Zenodo: https://zenodo.org/records/13799069

The **NLST-3D annotations**, adapted from slice-level bounding boxes, are provided within the shared codebase. The corresponding CT scans from the **National Lung Screening Trial (NLST)** can be requested through The Cancer Imaging Archive (TCIA): https://wiki.cancerimagingarchive.net/display/NLST

External validation datasets used in this study can be accessed from their official sources:

**LUNA16**: https://luna16.grand-challenge.org/Data/

**LUNA25**: https://luna25.grand-challenge.org/

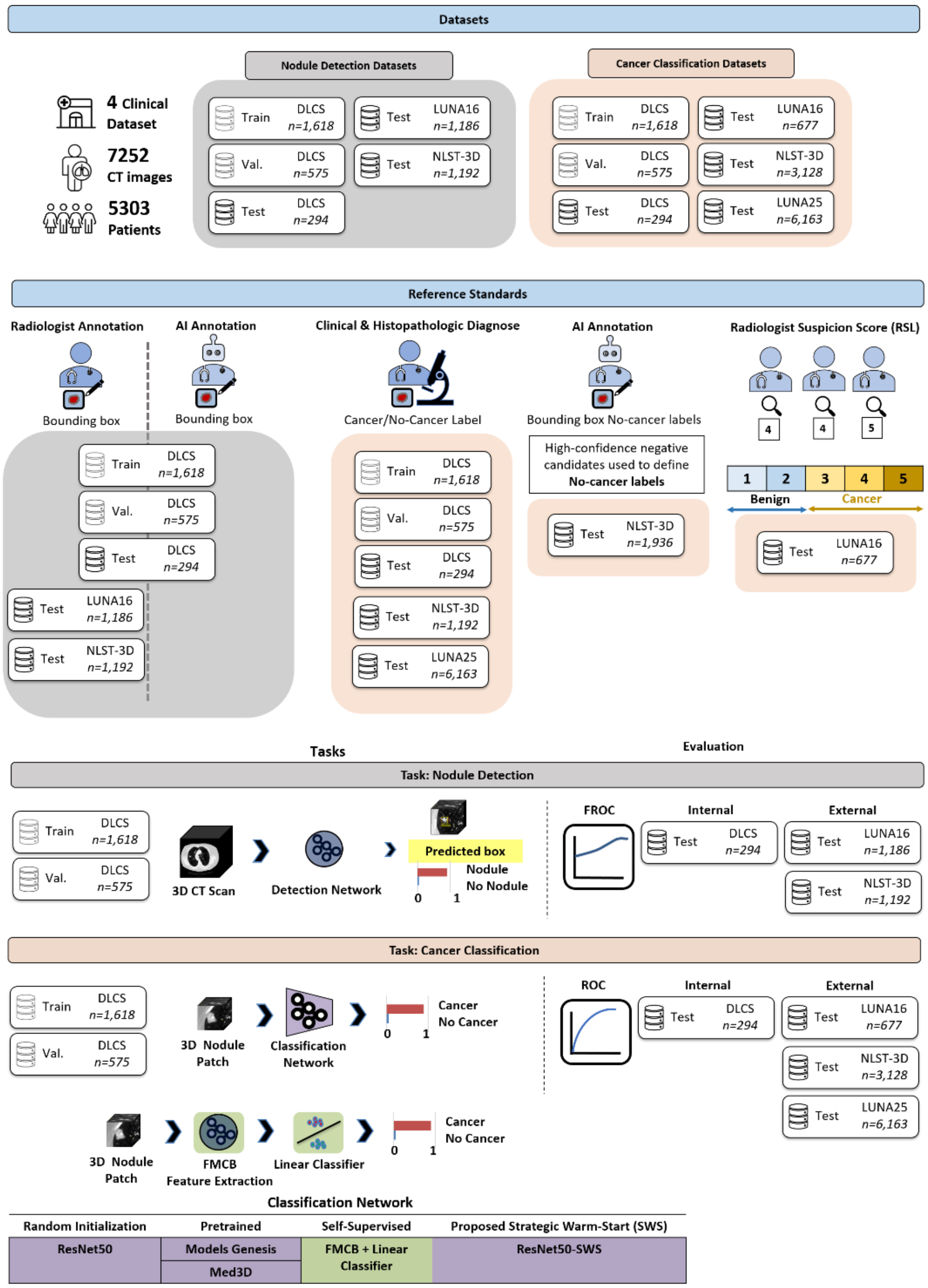

**Figure 1. Overview of dataset composition, reference standards, and benchmark design.** Datasets and task-specific splits used for lung nodule detection and cancer classification are shown (top). Reference standards differed across datasets, including pathology correlation, radiologist annotation, and high-confidence negative detection candidates used to define no-cancer labels (middle). Bottom panels illustrate task-specific model pipelines and evaluated approaches. DLCS served as the anchor dataset for training and validation, while all other datasets were used exclusively for external testing.

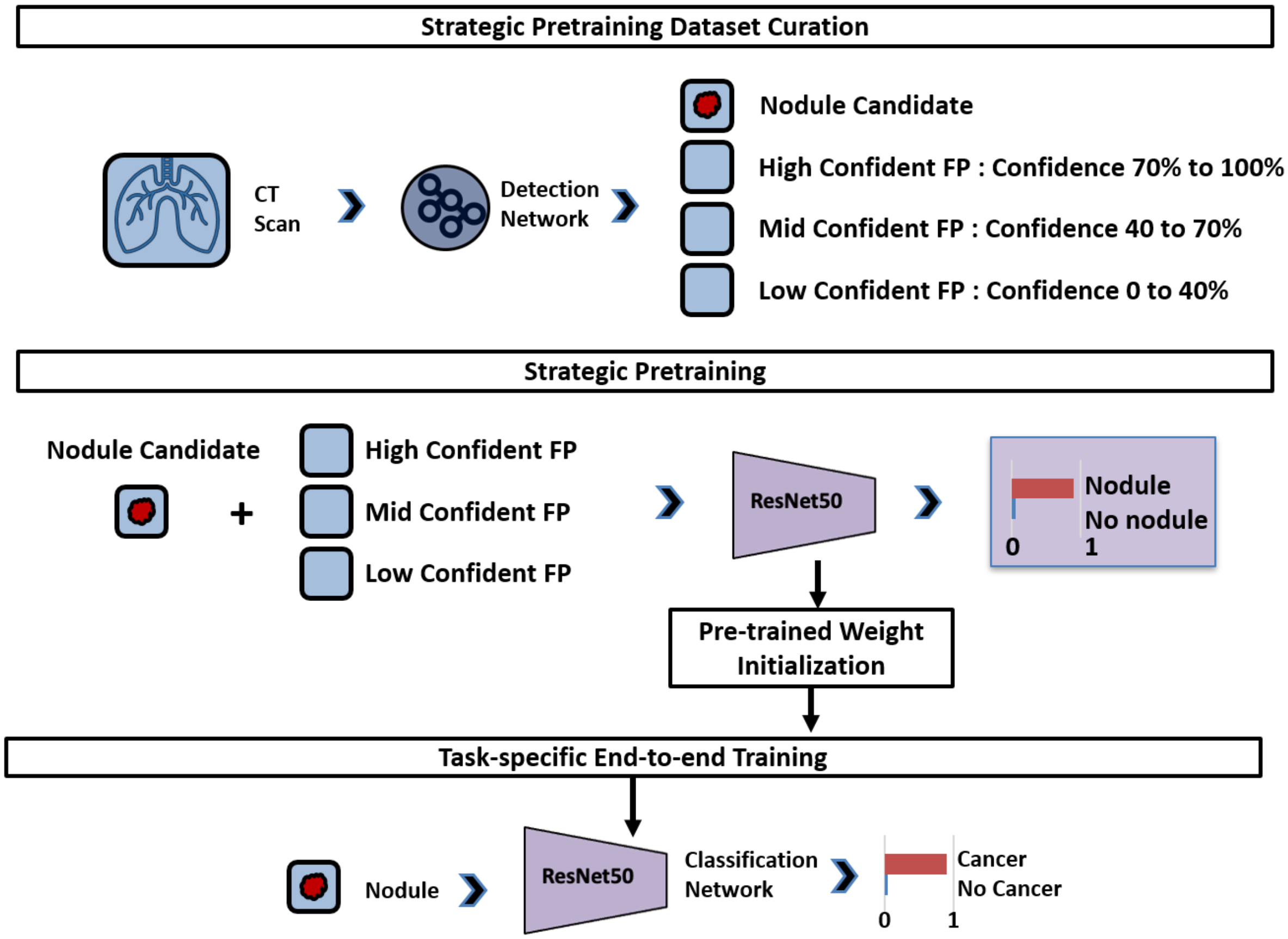

**Figure 2.** Overview of the Strategic Warm-Start (SWS) approach, illustrating (top) dataset curation for false positive, (middle) pretraining of ResNet50 on the curated dataset, and (bottom) transfer of pretrained weights for downstream cancer classification. ResNet50 was pretrained to distinguish 'nodule vs non-nodule' using detection-derived candidates before fine-tuning for 'cancer vs non-cancer.'

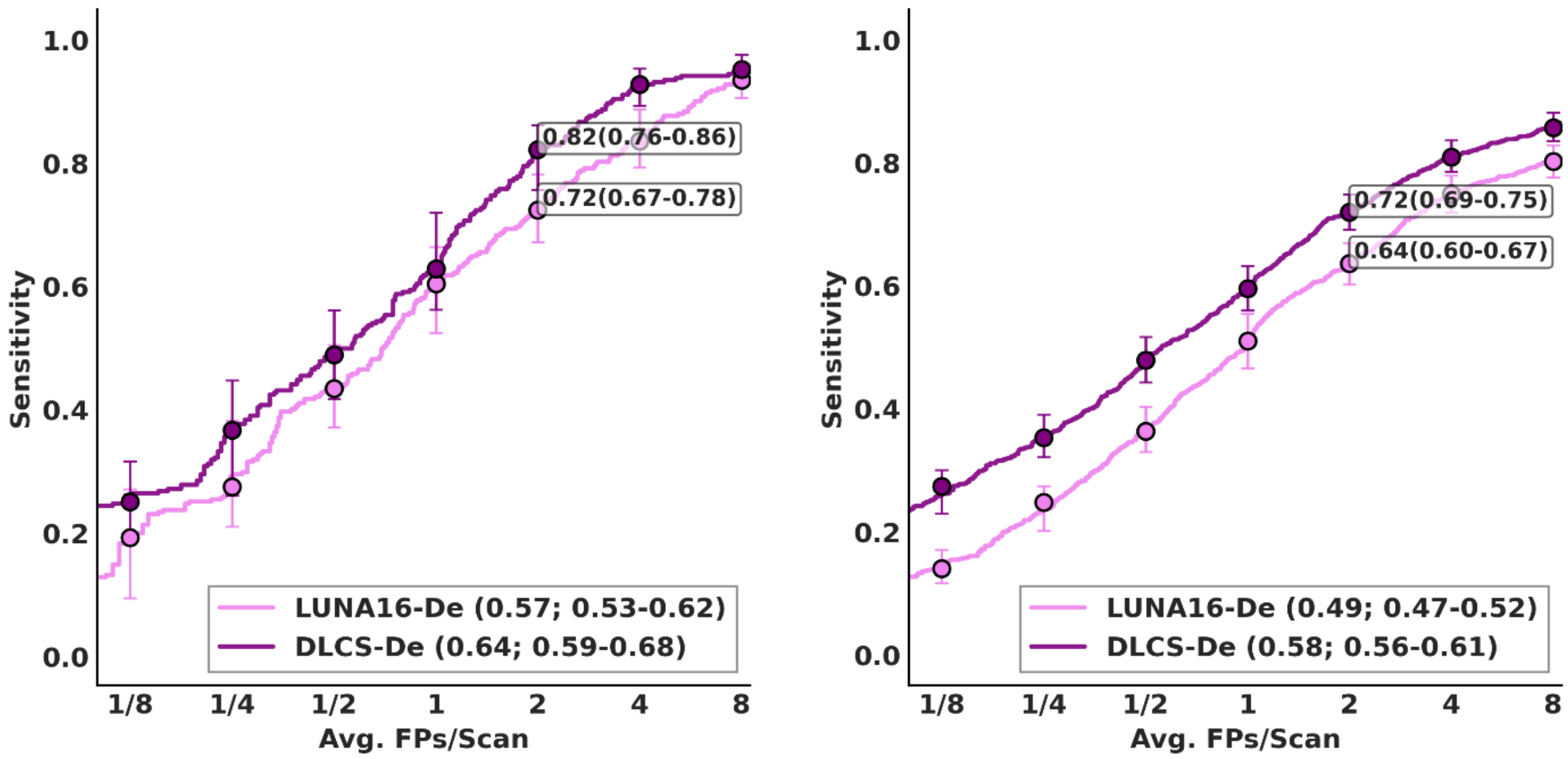

**(a) DLCS** **(b) NLST-3D**

**Figure 3.** Free-response receiver operating characteristic (FROC) performance on the internal test dataset (a) DLCS and external test dataset (b) NLST-3D comparing DLCS-De and LUNA16-De. Points denote sensitivity at predefined FP/scan operating points (1/8–8), with error bars indicating **95% bootstrap confidence intervals**. Legend reports **average sensitivity/** Competition Performance Metric **(CPM)** across the evaluated FP/scan range (1/8-8). Boxed values indicate sensitivity at 2 false positives per scan. CPM scores for each model are shown in parentheses in the legend.

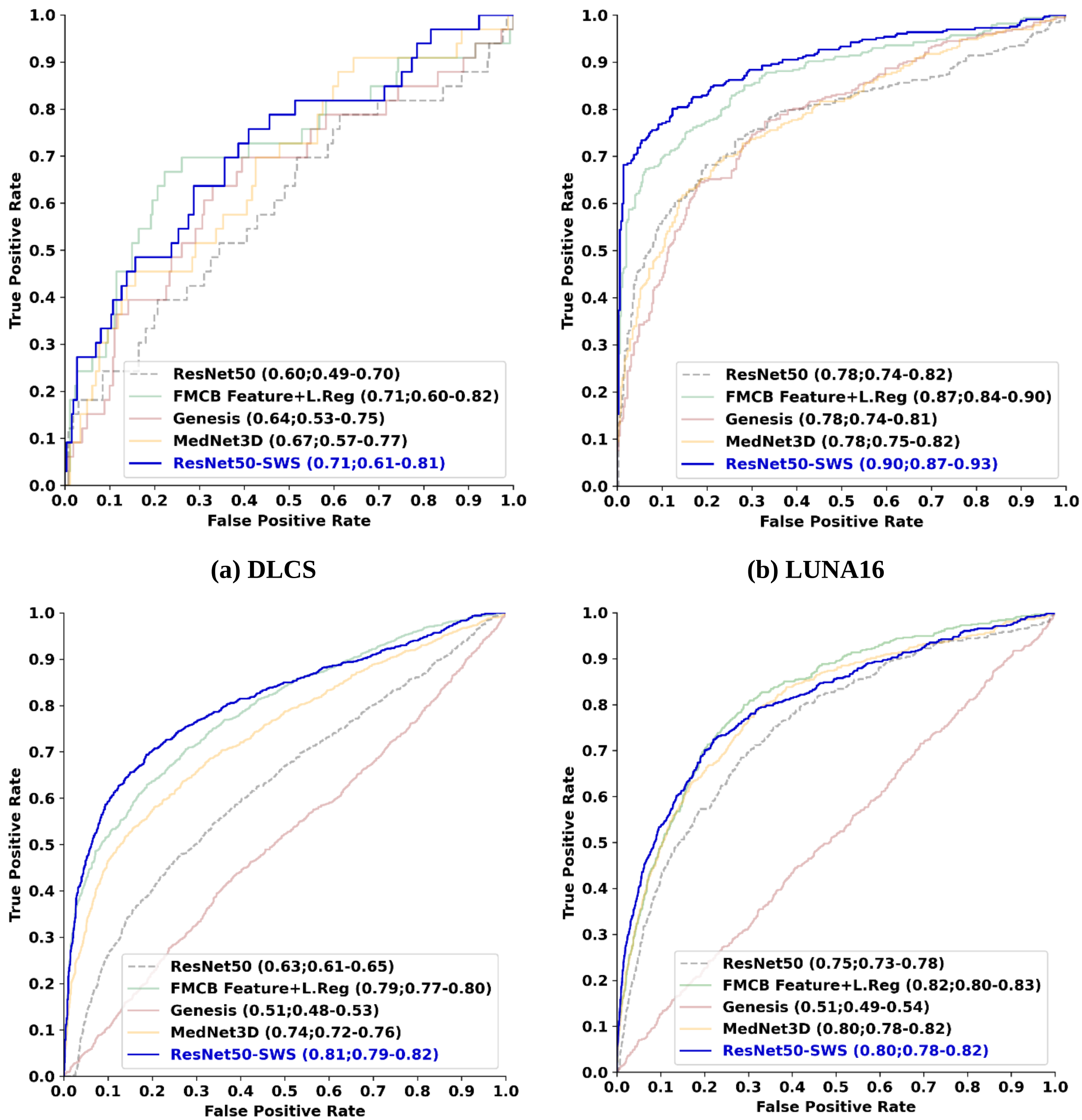

**(a) DLCS** **(b) LUNA16**

**(c) NLST-3D** **(d) LUNA25**

**Figure 4.** Nodule-level malignancy classification for 5 models trained on DLCS development set. Panels show receiver operating characteristic curves for testing on the following datasets: (A) DLCS internal validation, (B) LUNA16 external validation, (C) NLST-3D external validation, and (d) LUNA25 external validation. Values in parentheses indicate area under the curve and 95% confidence intervals.

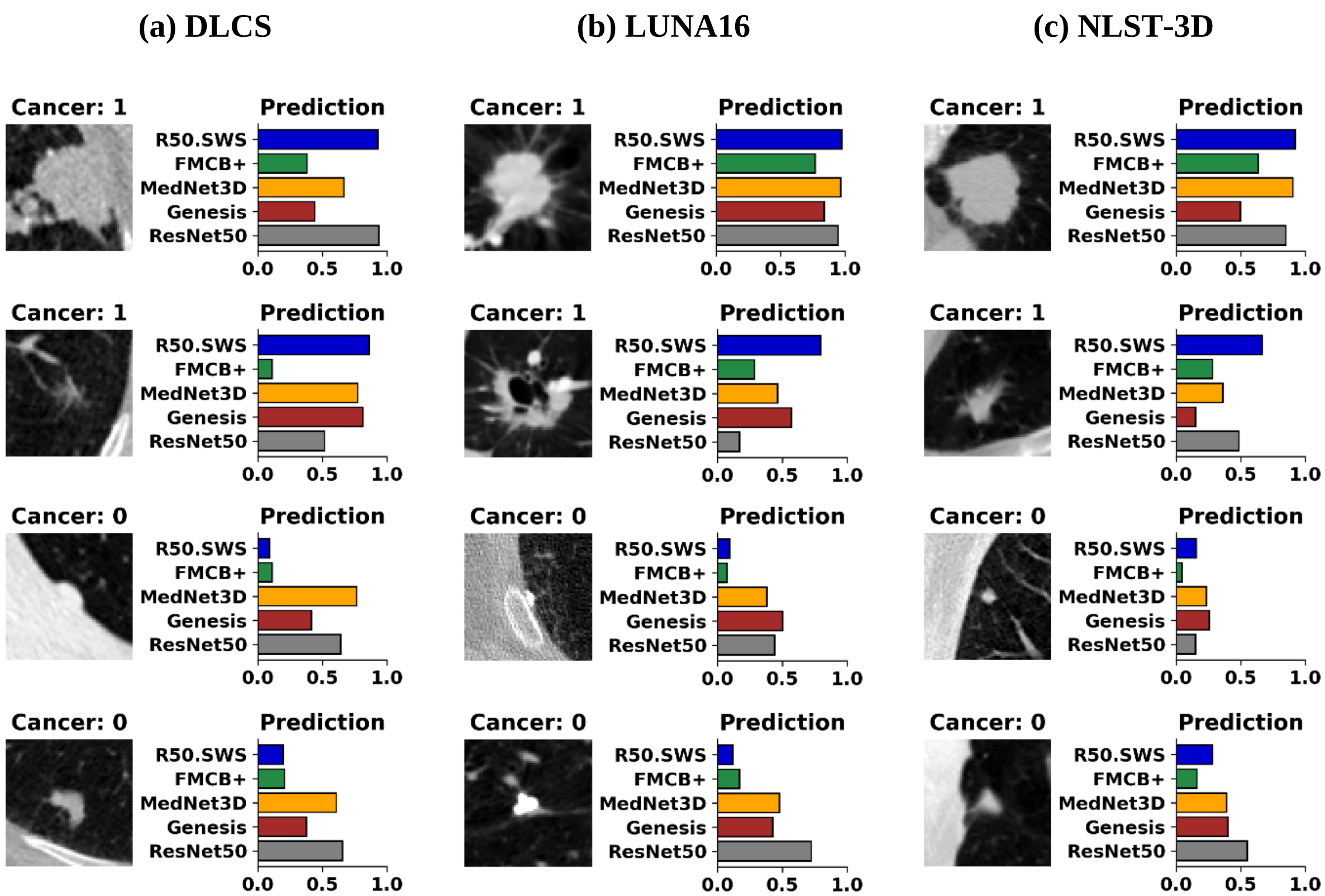

**Figure 5.** Examples of cancer classification results for (a) DLCS, (b) LUNA16, and (c) NLST-3D. Each image is derived from a 3D sub-volume patch and is labeled as cancer "1" or non-cancer "0" above the patch. Each patch is accompanied by histograms showing outputs from the 5 models. R50.SWS= ResNet50 Strategic Warm-Start (SWS); FMCB+= FMCB Features + logistic regression model.

Table 1. **Demographic distribution of the development and evaluation data dataset.**

| Dataset | Development and Validation | External Test | | |
|---|---|---|---|---|
| | Duke Lung Cancer Screening (DLCS) | National Lung Screening Trials (NLST) 3D | LUNA16 | LUNA25 |
| **Category** | | | | |
| **Patient** | | | | |
| **Patient (%)** | 1613 (100) | 969 (100) | 601 (100) | 2120 (100) |
| **CT Scans (%)** | 1613 (100) | 969 (100) | 601 (100) | 4069 (100) |
| **Gender (%)** | | | | |
| Male | 811 (50.3) | 572 (59.0) | Unknown | 1211 (57.1) |
| Female | 802 (49.7) | 397 (41.0) | Unknown | 909 (42.9) |
| **Age (years)** | | | | |
| Mean (min-max) | 66 (50-89) | 63 (55-74) | Unknown | 62 (55-76) |
| **Race (%)** | | | | |
| White | 1,195 (74.1) | 900 (92.9) | Unknown | 1975 (93.2) |
| Black/AA | 366 (22.7) | 43 (4.4) | Unknown | 76 (0.04) |
| Other/Unknown | 52 (3.2) | 26 (2.7) | Unknown | 69 (0.03) |
| **Ethnicity (%)** | | | | |
| Not Hispanic | 1,555 (96.4) | 954 (98.5) | Unknown | 2072 (97.7) |
| Unavailable | 52 (3.2) | 7 (0.7) | Unknown | 14 (0.01) |
| Hispanic | 6 (0.4) | 8 (0.8) | Unknown | 34 (2.0) |
| **Cancer (%)** | | | | |
| Benign | 1,469 (91.1) | 0 | Unknown | 1787 (84.3) |
| Malignant | 144 (8.9%) | 969 (100) | Unknown | 333 (15.7) |
| **Detection Task** | | | | |
| **Nodule Count* (%)** | **2487 (100)** | **1,192 (100)** | **1186 (100)** | **6163 (100)** |
| **Classification Task** | | | | |
| **Cancer (%)** | | | | |
| No cancer | 2,223 (89.4)† | 1936 (61.9)*** | 327 (48.3)** | 5608 (0.9)† |
| Cancer | 264 (10.6)† | 1,192 (38.1)† | 350 (51.7)** | 555 (0.1)† |

**Note:** ***Nodule-level counts; †= Histopathology & clinical follow-up; **** = Radiologist Suspicion Label (RSL); *****=AI annotated pseudo-labeled;

**Table 2. FROC sensitivity at the predefined false-positive (FP) per scan operating points of the LUNA16 challenge (1/8-8 FP/scan).** Average (CPM) denotes the mean sensitivity across these operating points, consistent with prior LUNA16 benchmark reporting.

| Model | 1/8 | 1/4 | 0.5 | 1.0 | 2.0 | 4.0 | 8.0 | Average (CPM) |
|---|---|---|---|---|---|---|---|---|
| **Liu et al. (2019) (11)** | 0.85 | 0.88 | 0.91 | 0.93 | 0.94 | 0.96 | 0.97 | 0.92 |
| **nnDetection (12)** | 0.81 | 0.89 | 0.93 | 0.95 | 0.97 | 0.98 | 0.99 | 0.93 |
| **LUNA16-De (14)** | 0.84 | 0.89 | 0.93 | 0.96 | 0.97 | 0.98 | 0.99 | 0.94 |
| **DLCS-De (our)** | 0.80 | 0.86 | 0.91 | 0.94 | 0.97 | 0.98 | 0.99 | 0.92 |

**Note: CPM =** Competition Performance Metric.

**Table 3: Model Performance (AUC) Across Datasets.**

| Model | DLCS (n=294) | LUNA16 (n=677) | NLST-3D (n=3128) | LUNA25 (n=6163) |
|---|---|---|---|---|
| **ResNet50** | 0.60 (0.49–0.70) | 0.78 (0.74–0.82)† | 0.63 (0.61–0.65)† | 0.75 (0.73–0.78)† |
| **FMBI** | 0.71 (0.60–0.82) | 0.87 (0.84–0.90)* | 0.79 (0.77–0.80)* | 0.82 (0.80–0.83) |
| **Genesis** | 0.64 (0.53–0.75) | 0.78 (0.74–0.81)† | 0.51 (0.48–0.53)† | 0.51 (0.49–0.54)† |
| **Med3D** | 0.67 (0.57–0.77) | 0.78 (0.75–0.82)† | 0.74 (0.72–0.76)† | 0.80 (0.78–0.82) |
| **ResNet50-SWS**** | 0.71 (0.61–0.81) | 0.90 (0.87–0.93) | 0.81 (0.79–0.82) | 0.80 (0.78–0.82) |

**Note**: Data are bootstrapped mean areas under the receiver operating characteristic curve (AUC), with 95% CIs in parentheses. Significance compared with the reference model (ResNet50-SWS) using the DeLong test is indicated by:* =($p < 0.05$), † =($p < 0.001$).** = Reference, n = number of nodules.

## Supplementary Materials

This supplement provides methodological and quantitative details that support the main manuscript and facilitate reproducibility. We include expanded cohort summaries and dataset split statistics (**Table S1**), along with paired bootstrap comparisons for detection performance across internal DLCS testing and external NLST-3D evaluation (**Table S2**). Additional implementation details (preprocessing, training configurations, and evaluation definitions) are provided to enable replication of the benchmark and fair comparison with future methods.

1. Supplementary Table S1. Demographic distribution of the data cohort used for training, development and test sets.
2. Supplement Table S2: Detection Performance on DLCS (Internal Test) and NLST-3D (External Test)

__Supplementary Table S1.__ **Demographic distribution of the data cohort used for training, development and test sets.** Note: N/A= not given.

| **Category** | | **All (%)** | **Training (%)** | **Validation (%)** | **Testing (%)** |
|---|---|---|---|---|---|
| **Duke Lung Cancer Screening Dataset** | | | | | |
| **Gender** | | | | | |
| | Male | 811 (50.28) | 559 (52.48) | 167 (46.78) | 85 (42.93) |
| | Female | 802 (49.72) | 499 (47.16) | 190 (53.22) | 113 (57.07) |
| | | | | | |
| **Age** | Mean (min-max) | 66 (50-89) | 66 (50-89) | 66 (55-78) | 66 (54-79) |
| | | | | | |
| **Race** | White | 1,195 (74.09) | 775 (73.25) | 280 (78.43) | 140 (70.71) |
| | Black/AA | 366 (22.69) | 247 (23.35) | 68 (19.05) | 51 (25.76) |
| | Other/Unknown | 52 (3.22) | 36 (3.40) | 9 (2.52) | 7 (3.54) |
| | | | | | |
| **Ethnicity** | | | | | |
| | Not Hispanic | 1,555 (96.40) | 1,019 (96.31) | 344 (96.36) | 192 (96.97) |
| | Unavailable | 52 (3.22) | 35 (3.31) | 12 (3.36) | 5 (2.53) |
| | Hispanic | 6 (0.37) | 4 (0.38) | 1 (0.28) | 1 (0.51) |
| | | | | | |
| **Smoking status** | | | | | |
| | Current | 826 (53.92) | 538 (53.48) | 189 (56.08) | 99 (52.38) |
| | Former | 704 (45.95) | 467 (46.42) | 147 (43.62) | 90 (47.62) |
| | Other/Unknown | 2 (0.13) | 1 (0.10) | 1 (0.30) | |
| **Cancer** | | | | | |
| | **Patient** | | | | |
| | Benign | 1,469 (91.07) | 965 (91.21) | 324 (90.76) | 180 (90.91) |
| | Malignant | 144 (8.93%) | 93 (8.79) | 33 (9.24) | 18 (9.09) |
| | | | | | |
| | Lung-RADS | | | | |
| | 1 | 8 (0.64) | 5 (0.61) | 2 (0.73) | 1 (0.64) |
| | 2 | 703 (56.20) | 463 (56.33) | 152 (55.68) | 88 (56.41) |
| | 3 | 219 (17.51) | 143 (17.40) | 49 (17.95) | 27 (17.31) |
| | 4A | 165 (13.19) | 106 (12.90) | 38 (13.92) | 21 (13.46) |
| | 4B | 113 (9.03) | 78 (9.49) | 21 (7.69) | 14 (8.97) |
| | 4X | 43 (3.44) | 27 (3.28) | 11 (4.03) | 5 (3.21) |
| | | | | | |
| | **Nodule** | | | | |
| | Benign | 2,223 (89.38) | 1,452 (89.74) | 510 (88.70) | 261 (88.78) |
| | Malignant | 264 (10.62) | 166 (10.26) | 65 (11.30) | 33 (11.22) |
| | | | | | |
| **Category** | | **All (%)** | **Training (%)** | **Validation (%)** | **Testing (%)** |
| **National Lung Screening Trial (NLST)** | | | | | |
| **Gender** | | | | | |
| | Male | 572 (59.03) | | | 572 (59.03) |
| | Female | 397 (40.97) | | | 397 (40.97) |
| **Age** | Mean (min-max) | 63 (55-74) | | | 63 (55-74) |
| **Race** | White | 900 (92.88) | | | 900 (92.88) |
| | Black/AA | 43 (4.44) | | | 43 (4.44) |
| | Other/Unknown | 26 (2.68) | | | 26 (2.68) |
| **Ethnicity** | | | | | |
| | Not Hispanic | 954 (98.45) | | | 954 (98.45) |
| | Unavailable | 7 (0.72) | | | 7 (0.72) |
| | Hispanic | 8 (0.83) | | | 8 (0.83) |
| **Smoking status** | | | | | |
| | Current | 535 (55.21) | | | 535 (55.21) |
| | Former | 434 (44.79) | | | 434 (44.79) |
| **Pack-year smoking history** | | | | | |
| | 21-30 years | 18 (1.86) | | | 18 (1.86) |
| | > 30+ years | 951 (98.14) | | | 951 (98.14) |
| **Study year of the last screening** | | | | | |
| | Year 0 | 265 (27.35) | | | 265 (27.35) |
| | Year 1 | 282 (29.10) | | | 282 (29.10) |
| | Year 2 | 422 (43.55) | | | 422 (43.55) |
| **Cancer** | | | | | |
| | **Patient** | | | | |
| | Malignant (Screen-detected) | 926 (95.56) | | | 926 (95.56) |
| | Malignant (Other) | 43 (4.44) | | | 43 (4.44) |
| | | | | | |
| | **Nodule** | | | | |
| | Malignant (Screen-detected) | 1,143 (95.89) | | | 1,143 (95.89) |
| | Malignant (Other) | 49 (4.11) | | | 49 (4.11) |
| | | | | | |
| | | | | | |
| | | | | | |

| **Category** | | **All (%)** | **Training (%)** | **Validation (%)** | **Testing (%)** |
|---|---|---|---|---|---|
| **LUNA16** | | | | | |
| | | | | | |
| **Gender** | N/A | | | | N/A |
| **Age** | N/A | | | | N/A |
| | | | | | |
| **Nodule Annotations** | Patients | 601 (100) | | | 601 (100) |
| | Nodule | 1186 (100) | | | 1186 (100) |
| | | | | | |
| **Radiologist Suspicion Label (RSL)** | | | | | |
| | **Nodule** | | | | |
| | Positive | 327 (48.3) | | | 327 (48.3) |
| | Negative | 350 (51.7) | | | 350 (51.7) |
| | | | | | |
| **LUNA25** | | | | | |
| | | | | | |
| **Gender** | | | | | |
| | Male | 1211 (57.12) | | | 1211 (57.12) |
| | Female | 909 (42.88) | | | 909 (42.88) |
| | | | | | |
| **Age** | Mean (min-max) | 62 (55-76) | | | 62 (55-76) |
| | | | | | |
| **Cancer Annotation** | | | | | |
| | **Nodules** | | | | |
| | Positive | 555 (0.09) | | | 555 (0.09) |
| | Negative | 5608 (0.91) | | | 5608 (0.91) |

**Supplement Table S2:** Detection Performance on DLCS (Internal Test) and NLST-3D (External Test)

| Dataset | Metric | LUNA16-De | DLCS24-De | Difference | P Value |
|---|---|---|---|---|---|
| **DLCS** | | *(n = 198)* | *(n = 198)* | *(DLCS24-De LUNA16-De)* | |
| **Internal test** | Avg sensitivity | 0.57 (0.53, 0.62) | 0.64 (0.59, 0.68) | 0.061 (0.031, 0.092) | < .001 |
| | Sensitivity @ 2 FP/scan | 0.72 (0.67, 0.78) | 0.82 (0.76, 0.86) | 0.099 (0.040, 0.141) | < .001 |
| **NLST-3D** | | *(n = 969)* | *(n = 969)* | | |
| **External** | Avg sensitivity | 0.49 (0.47, 0.52) | 0.58 (0.56, 0.61) | 0.093 (0.076, 0.106) | < .001 |
| | Sensitivity @ 2 FP/scan | 0.64 (0.60, 0.67) | 0.72 (0.69, 0.75) | 0.083 (0.064, 0.106) | < .001 |

**Note:** Data in parentheses are 95% CIs. Average sensitivity is calculated over 0.125-8 false positives (FP) per scan. Paired bootstrap comparisons were computed on scans common to both models within each dataset

The detection performance of the **DLCS24-De** and **LUNA16-De** models was compared across internal and external cohorts (Supplement Table S2). In the internal **DLCS** test set (*n* = 198), the DLCS24-De model demonstrated significantly higher average sensitivity (0.64 vs 0.57; *P* < .001) and sensitivity at 2 FP/scan (0.82 vs 0.72; *P* < .001) compared with LUNA16-De. This superiority was maintained during external validation on the **NLST-3D** cohort (*n* = 969), where DLCS24-De achieved an average sensitivity of 0.58 compared to 0.49 for LUNA16-De (difference, 0.093; 95% CI: 0.076, 0.106; *P* < .001). At the clinical threshold of 2 FP/scan, DLCS24-De reached a sensitivity of 0.72, representing a statistically significant improvement over the 0.64 achieved by the LUNA16-trained detector (*P* < .001). These results indicate that the DLCS-trained detector provides superior generalizability for screening-detected nodules compared with models trained on the more traditional LUNA16 dataset.